\documentclass{article}
\usepackage{mathrsfs,pifont,spconf,amsmath,graphicx,hyperref,amsfonts,amssymb,units,hyperref,multirow,marvosym,booktabs,xcolor}

\newcommand{\yes}{\ding{51}}
\newcommand{\no}{~}

\title{Multi-View Crowd Counting with Self-Supervised Learning}
\name{$^{1}~$Hong Mo ~~ $^{2}~$Xiong Zhang ~~ $^{1}~$Tengfei Shi ~~ $^{1}~$Zhongbo Wu}
  \address{$^{1}$ Hubei University of Arts and Science \\
   $^{2}$ Neolix Autonomous Vehicle
  }

\begin{document}

\maketitle
\begin{abstract}
  \label{sec:abs}
Multi-view counting (MVC) methods have attracted significant research attention and stimulated remarkable progress in recent years.
Despite their success, most MVC methods have focused on improving performance by following the fully supervised learning (FSL) paradigm, which often requires large amounts of annotated data.
In this work, we propose {\bf SSLCounter}, a novel self-supervised learning (SSL) framework for MVC that leverages neural volumetric rendering to alleviate the reliance on large-scale annotated datasets.
SSLCounter learns an implicit representation w.r.t. the scene, enabling the reconstruction of continuous geometry shape and the complex, view-dependent appearance of their 2D projections via differential neural rendering.
Owing to its inherent flexibility, the key idea of our method can be seamlessly integrated into exsiting frameworks.
Notably, extensive experiments demonstrate that SSLCounter not only demonstrates state-of-the-art performances but also delivers competitive performance with only using 70\% proportion of training data, showcasing its superior data efficiency across multiple MVC benchmarks.

\end{abstract}

\section{Introduction}
\label{sec:intro}

Multi-view counting (MVC) approaches have demonstrated substantial effectiveness and achieved notable results on well-established benchmarks \cite{zhang2019wide,zhang20203d,zhang2021cross,mo2024countformer}. 
The foundational pipeline in prevailing methods initiates by transforming multi-view (MV) features that extracted from MV images into a scene-level volumetric representation,
and subsequently deriving the volume density through the application of convolutional layers to the resultant volumetric features \cite{mo2024countformer}. 


Despite its significant potential, contemporary research predominantly follows the fully supervised learning (FSL) paradigm in pursuit of higher performance \cite{mo2024countformer,zhang20223d}, i.e., employing a loss function that minimizes the discrepancy between the ground-truth volume density and its predicted counterpart, which require large-scale datasets with precise annotations.
However, obtaining such annotations is labor-intensive and costly \cite{zhang2019wide,zhang2021cross,ristani2016performance,ferryman2009pets2009}, e.g., maintaining cross-view consistence for each head-point proves time-consuming,
which consequently limits their practical deployment in real-world scenarios.



In this work, we build SSLCounter by integrating the self-supervised learning (SSL) strategy into existing MVC approaches,
inducing the networks to achieve robust representations learning from unlabeled data, thereby significantly reducing their reliance on large-scale annotated datasets and mitigating the data scarcity problem.
The key design philosophy of SSLCounter is chiefly concerned with enhancing the representation capability of scene-level volumetric features that derived from MV images, 
particularly by improving the model's capacity for fine-grained, cross-view contextual capture and scene-level geometric structure understanding.

\begin{figure*}[htbp] 
    \centering
    \includegraphics[width=0.9\textwidth]{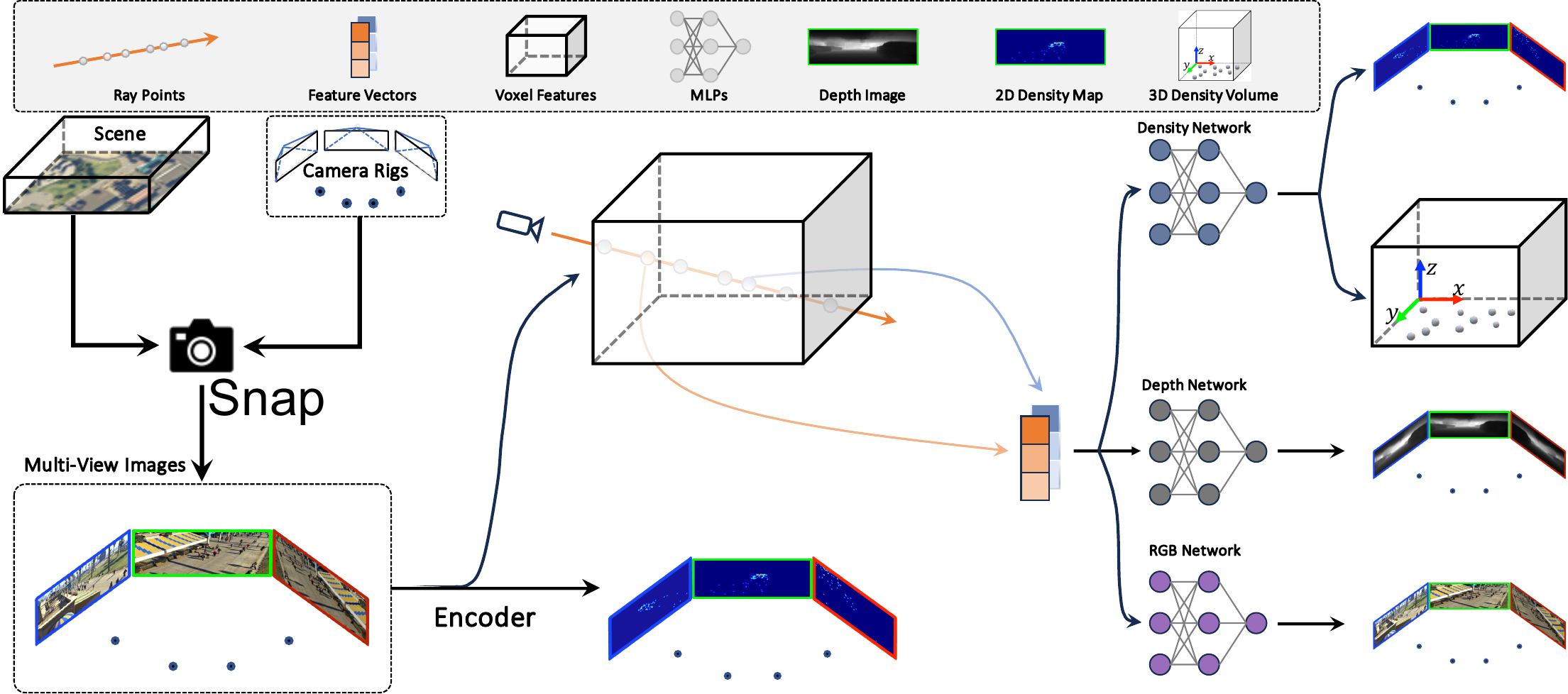} 
    \vspace{-10pt}
    \caption{
      {\bf Framework of the SSLCounter.}
      SSLCounter processes multi-view images through an encoder to acquire the scene-level volume representation and image-level density maps, leveraging the volumetric neural rendering to reconstruct the RGB image, produce the depth map, and derive both the image-level density map and the scene-level density volume.
    }
    \label{fig:arch}
\end{figure*}

To achieve this goal, as in Figure.~\ref{fig:arch}, MV image-level features are exploited to acquire a coherent scene-level volume representation while simultaneously estimating the image-level density maps.
Subsequently, the differentiable volumetric rendering \cite{chibane2020neural,mildenhall2020nerf} is employed to facilitate the reconstruction of the geometric representation and the appearance of various 2D projections.
For instance, a perspective-aware depth map can be estimated by applying ray tracing to the volume representation, 
the corresponding image-level density map can be derived by integrating the scene-level volume density along each ray, 
and the RGB image can be rendered by rendering the point-wise colors w.r.t. points that constitute the ray.

After acquiring the depth map, reconstructed RGB image, 3D scene-level density volume, and rendered image-level density map, 
critical supervision beyond the density volume discrepancy loss can be employed to further regularize and improve the model. 
Specifically, the rendered density map can be supervised by enforcing consistency with the prediction from the encoder, which effectively anchors the SSL paradigm in a reliable feature space.
Additionally, the depth map can be regularized by encouraging alignment with the depth prior, thereby facilitating the recovery of geometrically consistent 3D structures.
Furthermore, the RGB reconstruction process is fully differentiable, which inherently enables and strengthens the entire SSL framework.

In summary, SSLCounter is a highly versatile module that maintains full compatibility with established MVC pipelines, 
and extensive experiments confirm that such design effectively addresses the critical challenge of data scarcity, offering a practical solution for real-world MVC applications.

\section{Methodology}
As illustrated in Figure.~\ref{fig:arch}, 
SSLCounter comprises a encoder that simultaneously derives the scene-level volumetric representation as well as estimates the image-level density map, 
and a differentiable neural decoder to docode the volumedtric features into the image-level density map, scene-level density volume, depth map, and RGB image.

\subsection{Encoder}
The encoder takes multi-view (MV) images 
as input, extracting MV pyramid features for each view, lifting the features to the scene-level volume representation as well as performing the image-level density estimation.
In this work, we adopt a structure similar to CountFormer~\cite{mo2024countformer}, 
due to its highly versatility and scalability \footnote{~additional formulations  \cite{zhang20223d,philion2020lift,liu2023sparsebev}  remain available},
 to derive the volume representation $\mathbf{V}$ and the image-level density map.

\subsection{Decoder}
The decoder samples $\rm {K}$ rays $\rm R = \{\rm \mathbf  r_{ i}\}_{ i \le  K}$ from various viewing directions and adopts differentiable volume rendering to decode the color, depth, and density for each ray, which may further facilitates the incorporation of priors into the acquired image features via supplementary rendering supervision.
To reach this target, the signed distance field (SDF)~\cite{wang2021neus,oechsle2021unisurf} is utilized to represent the geometric structure of the scene, 
since it is capable of encoding high-quality details.

Specifically, for each ray $ \rm \mathbf r$, the subscript $\rm i$ is omitted to keep clarity, with camera origin $\rm \mathbf o$ and view direction $\rm \mathbf d$, $\rm M$ points $\{\rm  \mathbf p_i=\mathbf o + t_i \cdot \mathbf d\}_{ i \le M}$ that belong the ray are randomly sampled, where $\rm \mathbf p_i$ represents the scene-level coordinates of the sampled points, and $\rm t_j$ denotes the corresponding depth along $\rm r$.
For each ray point $\rm \mathbf p_i$, a feature embedding vector $\rm \mathbf f_i$ is extracted from the volumetric representation $\mathbf{V}$ through trilinear interpolation, 
and the SDF field $\rm s_i$ is predicted with a multi-layer perceptrons (MLPs) $\rm \phi_{SDF}(\mathbf{p}_i,\mathbf{f}_i)$.
Similarly, the color field $\rm c_i$ can also be parameterized with MLPs $\rm \phi_{RGB}(\mathbf{p}_i,\mathbf{f}_i,\mathbf{d})$, and the scene-level volume density field $\rm d_i$ could be estimated with MLPs $\rm \phi_{Density}(\mathbf{p}_i,\mathbf{f}_i)$.
Consequently, the depth $\rm {Z}_r$, the RGB value $\rm {C}_r$, and the image-level 2D density $\rm {D}_r$ are calculated by integrating the corresponding predicted fields along that ray
\begin{align}
\rm {\mathbf Z}_r=\sum_{i \le M} w_i \cdot t_i; ~
\rm{\mathbf C}_r=\sum_{i \le M} w_i \cdot c_i; ~
\rm{\mathbf D}_r=\sum_{i \le M} w_i \cdot d_i,
\end{align}
where $\rm w_i$ weights the occlusion as $\rm w_i=\prod_{\rm k\le i-1}(1-\alpha_k)\cdot\alpha_i$, and $\rm \alpha_k$ approximates the opacity value with 
\begin{align}
  \rm \alpha_k=\rm {\mathrm{max}\Big (\nicefrac{\rm \big(\delta(s_k)-\delta(s_{k+1})\big)}{\delta(s_k)}, 0\Big)},
\end{align}
where $\delta(\cdot)$ is the sigmoid function that defined as $\rm \delta(s)=\nicefrac{1.0}{\big(1+\exp(-s\cdot\beta)\big)}$ with learnable parameter $\beta$.

\subsection{SSL}
Besides using the conventional mean squared error (MSE) loss to supervise the estimated density map and density volume against their ground-truth counterparts, 
one may further apply the SSL strategy to regularize the model. 

Firstly, the estimated depth $\rm{\mathbf Z}$ should be consistent with the depth prior $\rm \bar{\mathbf Z}$, which can be derived from off-the-shelf tool \footnote{\url{https://depth-anything.github.io/}}, 
facilitating the reconstruction of continuous 3D geometric structure from MV images.
Secondly, the rendered density map $\rm {\mathbf D}$ should be consistent with that predicted by the encoder $\rm \bar{\mathbf D}$, which encourages the learning of a spatial meaningful density volume. 
Finally, for the RGB reconstruction, a reconstruction loss can be employed to encourage the rendered RGB color $\rm {\mathbf C}$ to match the ground-truth $\rm \bar{\mathbf C}$, improving the representation capability of the volume representation $\mathbf{V}$.

Consequently, the overall optimization objective can be formulated as
$\mathbf{L}_{\text{SSLCounter}} = \mathbf L_{\text{FSL}} + \mathbf L_{\text{SSL}}$ 
, where the fully supervised loss $\mathbf L_{\text{FSL}}$=$\rm \mathop{MSE(\bar{\mathbf D}, \hat{\mathbf D})}$ + $\rm \sum_{\mathbf{r} \in R}\sum_{i \le M}\mathop{MSE}(\rm d_i, \bar{d_i}) $ and 
the self-supervised regularizer 
$\mathbf L_{\text{SSL}}$=$\rm \sum_{r \in R}\mathop{MSE}(\mathbf{D}_r, \bar{\mathbf D}_r)$ + $\rm \sum_{r \in R}\mathop{MSE}(\mathbf Z_r, \bar{\mathbf Z}_r) + \sum_{r \in R}\mathop{MSE}(\mathbf C_r, \bar{\mathbf C}_r)$.
This joint optimization enables the model to leverage both labeled and unlabeled data, improving generalization and reducing reliance on large-scale annotations.

\section{Experiments}

\subsection{Quantitative Results}
We quantitative evaluate the effectiveness of SSLCounter on the CityStreet \cite{zhang2019wide}, PETS2009 \cite{ferryman2009pets2009}, and the CVCS \cite{zhang2021cross} benchmarks.
Following conventional \cite{zhang20223d,zhang2022calibration,zhang2022wide,mo2024countformer}, the mean absolute error (MAE$~\downarrow$ ) and normalized mean absolute error (NAE$~\downarrow$) are employed to quantify the performances.

As Table \ref{tb:pets2} presents the comparison results on the PETS2009 dataset, 
SSLCounter demonstrates clear superiority over most state-of-the-art (SOTA) approaches 
\cite{zhang2022wide,zhang20223d,zhang2021cross,zheng2021learning,zhang2022calibration,zhai2022co} and achieves comparable performance with CountFormer~\cite{mo2024countformer}, which is reasonable considering
the saturated performance on this dataset.
In contrast, the CityStreet dataset \cite{zhang2019wide} contains a larger crowd distribution, severe dynamic occlusions from the environment, and diverse scale variations caused by perspective projection,
Table \ref{tb:CityStreet_results} illustrates that our method is superior to those proposed by \cite{zhang2022wide,zhang20223d,zhang2021cross,zhang2022calibration,zhai2022co,mo2024countformer} in terms of MAE/NAE for both scene-level and single-view level, maintaining a
SOTA performance.
Similarity, on the most challenge benchmark CVCS \cite{zhang2021cross}, 
Table \ref{tb:cvcs_results} demonstrates that our approach achieves an impressive MAE/NAE and sets a new SOTA performance over all current works \cite{zhang2021cross,zhang2019wide,zhang2022calibration,zhang20223d,mo2024countformer}.
As anticipated, the SSL paradigm not only strengthens the model's learning capacity and feature representation ability but also advances its comprehension of geometric structure, resulting in superior overall performance.

\begin{table}[htbp]
  \centering
  \resizebox{1.0\columnwidth}{!}
  {
  \begin{tabular}{l|l|lll}
  \toprule
  \multicolumn{1}{c|}{Method}            & \multicolumn{1}{|c|}{Scene}    & \multicolumn{1}{c}{$\mathrm{C}_1$}         & \multicolumn{1}{c}{$\mathrm{C}_2$}        & \multicolumn{1}{c}{$\mathrm{C}_3$}        \\
  \hline
      MVMS \cite{zhang2019wide}      &   ~~3.49/0.124~   & ~1.66/0.084~ & ~2.58/0.103~  & ~3.46/0.127~ 	\\
      MVMSR \cite{zhang2022wide}     &   ~~3.62/0.130  & ~1.57/0.077 & ~2.38/0.097 & ~3.64/0.133 	\\
      3D Counting  \cite{zhang20203d} &   ~~3.15/~~ -   & \multicolumn{1}{c}{-}& \multicolumn{1}{c}{-} & \multicolumn{1}{c}{-} 	\\
      3D Counting \cite{zhang20223d} &   ~~3.20/~~ -   & \multicolumn{1}{c}{-}& \multicolumn{1}{c}{-} & \multicolumn{1}{c}{-} 	\\
      CVCS \cite{zhang2021cross} &   ~~5.17/0.165   & \multicolumn{1}{c}{-}& \multicolumn{1}{c}{-} & \multicolumn{1}{c}{-} 	\\
      CVF \cite{zheng2021learning} &   ~~3.08/~~ -   & ~1.66/~~ - & ~2.36/~~ - & ~3.41/~~ - 	\\
      FCLs \cite{qiu2019cross} &   ~~3.40/~~ -   & ~1.77/~~ - & ~2.74/~~ - & ~3.56/~~ - 	\\
      CF-MVCC-C \cite{zhang2022calibration} &   ~~3.84/0.125   & \multicolumn{1}{c}{-}& \multicolumn{1}{c}{-} & \multicolumn{1}{c}{-} 	\\
      CF-MVCC \cite{zhang2022calibration} &   ~~3.46/0.116   & \multicolumn{1}{c}{-}& \multicolumn{1}{c}{-} & \multicolumn{1}{c}{-} 	\\
      CoCo-GCN \cite{zhai2022co} &   ~~2.97/0.109 -   & ~1.36/~~ - & ~1.54/~~ - & ~2.25/~~ - 	\\
      CountFormer \cite{mo2024countformer} &   ~~0.74/0.030   & ~0.42/0.031 &~0.65/0.037 & ~0.84/0.036 	\\
  \hline
  SSLCounter &   ~~1.13/0.041   & ~0.51/0.039 &~1.02/0.072 & ~1.74/0.083 	\\
  \bottomrule
  \end{tabular}
  }
  \vspace{-11pt}
  \caption{ {\bf Performance on PETS2009.}
  The table exhibits the scene-level and camera-view counting performances using the mean absolute error and the relative mean absolute error (MAE $\downarrow$ / NAE $\downarrow$) on the PETS2009 \cite{ferryman2009pets2009} dataset.
  }
  \label{tb:pets2}
\end{table}

\begin{table}
  \centering
  \resizebox{1.0\columnwidth}{!}
  {
  \begin{tabular}{l|l|lll}
  \toprule
      \multicolumn{1}{c}{Method} & \multicolumn{1}{|c|}{Scene}    & \multicolumn{1}{c}{$\mathrm{C}_1$}         & \multicolumn{1}{c}{$\mathrm{C}_3$}        & \multicolumn{1}{c}{$\mathrm{C}_4$}        \\
  \hline
      MVMS \cite{zhang2019wide}      &   ~~7.36/0.096~   & ~7.87/0.110~ & ~6.87/0.100~  & ~6.24/0.081~ 	\\
      MVMSR \cite{zhang2022wide}     &   ~~6.98/0.086  & ~7.05/0.094& ~6.49/0.089 & ~6.44/0.078 	\\
      3D Counting  \cite{zhang20203d} &   ~~7.54/~~ -   & \multicolumn{1}{c}{-}& \multicolumn{1}{c}{-} & \multicolumn{1}{c}{-} 	\\
      3D Counting  \cite{zhang20223d} &   ~~7.12/0.091   & \multicolumn{1}{c}{-}& \multicolumn{1}{c}{-} & \multicolumn{1}{c}{-} 	\\
      CVCS \cite{zhang2021cross} &   ~~9.58/0.117   & \multicolumn{1}{c}{-}& \multicolumn{1}{c}{-} & \multicolumn{1}{c}{-} 	\\
      CVF \cite{zheng2021learning} &   ~~7.08/~~ -   & ~7.38/~~ - & ~6.85/~~ - & ~5.18/~~ - 	\\
      FCLs \cite{qiu2019cross} &   ~~7.71/~~ -   & ~7.67/~~ - & ~7.49/~~ - & ~5.50/~~ - 	\\
      CF-MVCC-C \cite{zhang2022calibration} &   ~~8.06/0.102   & \multicolumn{1}{c}{-}& \multicolumn{1}{c}{-} & \multicolumn{1}{c}{-} 	\\
      CF-MVCC \cite{zhang2022calibration} &   ~~8.24/0.103   & \multicolumn{1}{c}{-}& \multicolumn{1}{c}{-} & \multicolumn{1}{c}{-} 	\\
      CoCo-GCN \cite{zhai2022co} &   ~~6.19/0.084 -   & ~6.41/~~ - & ~6.79/~~ - & ~4.93/~~ - 	\\
      CountFormer \cite{mo2024countformer} &   ~~4.72/0.058   & ~5.26/0.079& ~5.92/0.081  & ~4.21/0.069 	\\
  \hline
  SSLCounter &   ~~3.97/0.051   & ~4.87/0.071& ~5.14/0.080  & ~3.57/0.052 	\\
  \bottomrule
  \end{tabular}
  }
  \vspace{-11pt}
  \caption{ {\bf Performance on CityStreet.}
  The table shows the scene-level and camera-view counting performances using the mean absolute error and the relative mean absolute error ( MAE $\downarrow$ / NAE $\downarrow$) on the CityStreet \cite{zhang2019wide} dataset.
  }
  \label{tb:CityStreet_results}
\end{table}

\begin{table}
  \centering
  \resizebox{0.52\linewidth}{!}
  {
  \begin{tabular}{l|cc}
  \toprule[1pt]
      \multicolumn{1}{c|}{Method} &~$\mathrm{MAE}$~&~$\mathrm{NAE}$~~\\
  \hline
      CVCS  \cite{zhang2021cross} &   7.22 & 0.062 \\
      MVMS \cite{zhang2019wide} &   9.30 & 0.080 \\
      CF-MVCC \cite{zhang2022calibration} &   16.5 & 0.140 \\
      CF-MVCC-C \cite{zhang2022calibration} &   13.9 & 0.118 \\
      3D Counting \cite{zhang20223d} & 13.3 & 0.123 \\
      CountFormer \cite{mo2024countformer} &  4.79 & 0.039 \\
  \hline
  SSLCounter &  4.01 & 0.032 \\
  \bottomrule
  \end{tabular}
  }
  \vspace{-7pt}
  \caption{ {\bf Performance on CVCS.}
  The table illustrates the scene-level counting performance using the mean absolute error (MAE $\downarrow$) and the relative mean absolute error (NAE $\downarrow$) on the CVCS \cite{zhang2021cross} dataset, which stands for the most challenging MVC benchmark.
  }

  \label{tb:cvcs_results}
\end{table}
\begin{table}
\centering
\resizebox{0.9\columnwidth}{!}
{%
\begin{tabular}{c |c c |c c}
\toprule
\multicolumn{1}{c|}{\multirow{2}{*}{\% Training}} & \multicolumn{2}{c|}{~~~~SSL~~~~} & \multicolumn{2}{c}{Performance}\\\cline{2-5}
& Training & Validation & MAE ($\downarrow$) & NAE ($\downarrow$)\\
\toprule

\multirow{2}{*}{70\%}   
& \no &  \no & 7.17& 0.095 \\
& \yes & \no & 5.48 & 0.071 \\
\hline
\multirow{3}{*}{100\%}   
& \no & \no &4.72& 0.058 \\
& \yes & \no & 3.97 & 0.051 \\
& \yes & \yes & 2.81 & 0.034 \\
\bottomrule
\end{tabular}
}
\vspace{-5pt}
\caption{\textbf{Ablation Study on CityStreet.} 
The table presents the ablation results on the CityStreet \cite{zhang2019wide} dataset using scene-level MAE $\downarrow$ ane NAE $\downarrow$, where \% Training refers the proportion of data used for training, and SSL indicates applying the SSL on the corresponding data part.
}
\label{tb:ab_citystreet}
\end{table}

\subsection{Qualitative Results}
\begin{figure} 
    \centering
    \includegraphics[width=0.97\columnwidth]{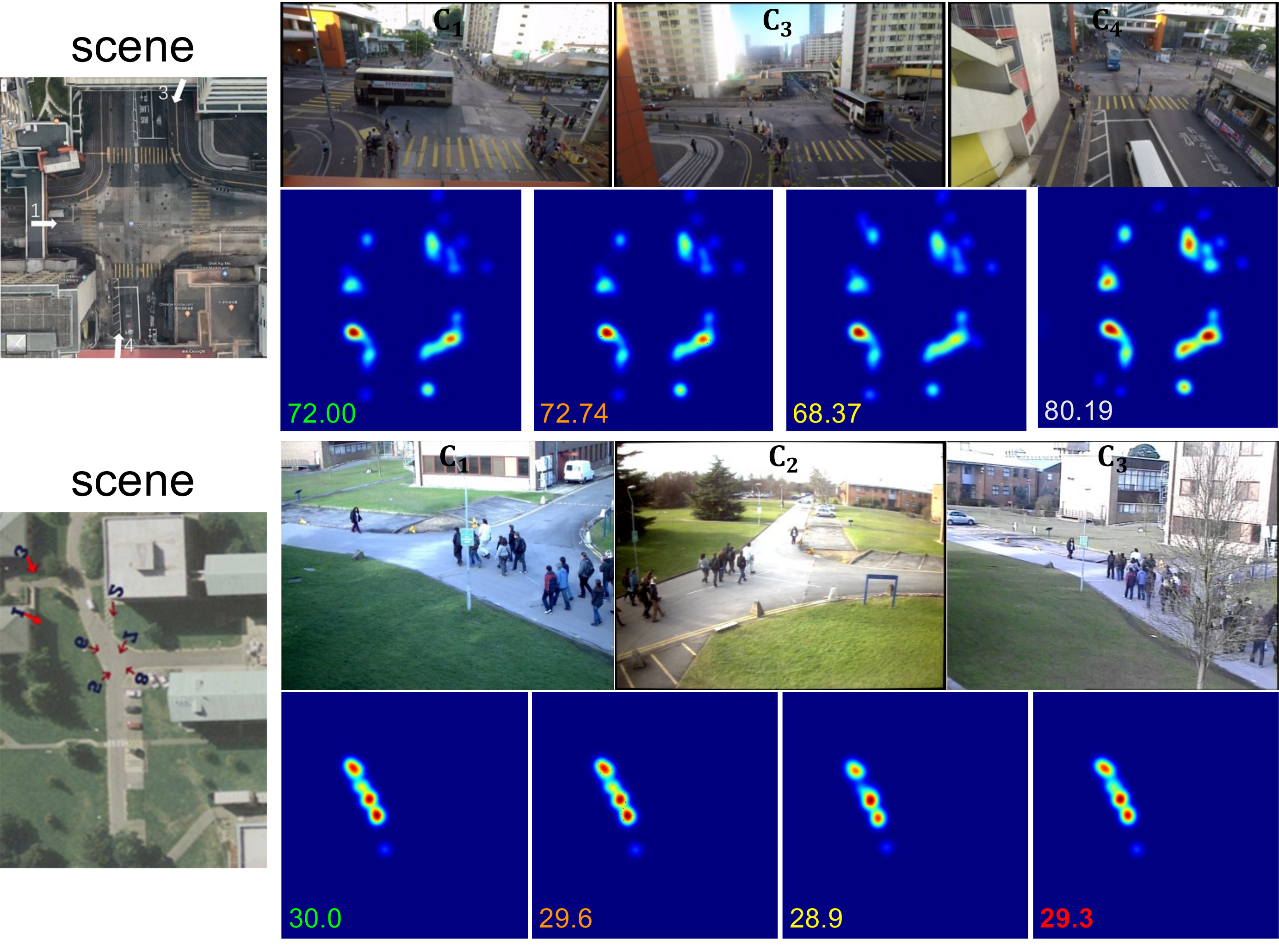} 
  \vspace{-2pt}
    \caption{{\bf Qualitative Results.} 
  The figure exhibits several typical scenarios on the CityStreet (with 3 views) and PETS2009 (with 3 views) datasets, including occlusion and congested crowds.   
  For each sample, the multi-view images, the {\color{green} ground truth} scene-level density and estimated density from {\color{orange} our SSLCounter method}, {\color{yellow}3D Counting approach}\cite{zhang20203d}, and {\color{red}the CVCS} \cite{zhang2021cross} are presented in the bird's eye view.}
    \label{fig:cvcs_2009}
\end{figure}

To better evaluate the robustness of SSLCounter in challenging scenarios, we selected representative samples from the CityStreet and PETS2009 testing sets, as shown in Figure \ref{fig:cvcs_2009}, featuring severe occlusion and highly congested crowds. 
Qualitative experimental results confirm the efficacy of our approach in handling occluded instances.

\begin{figure}
    \centering
    \includegraphics[width=0.97\columnwidth]{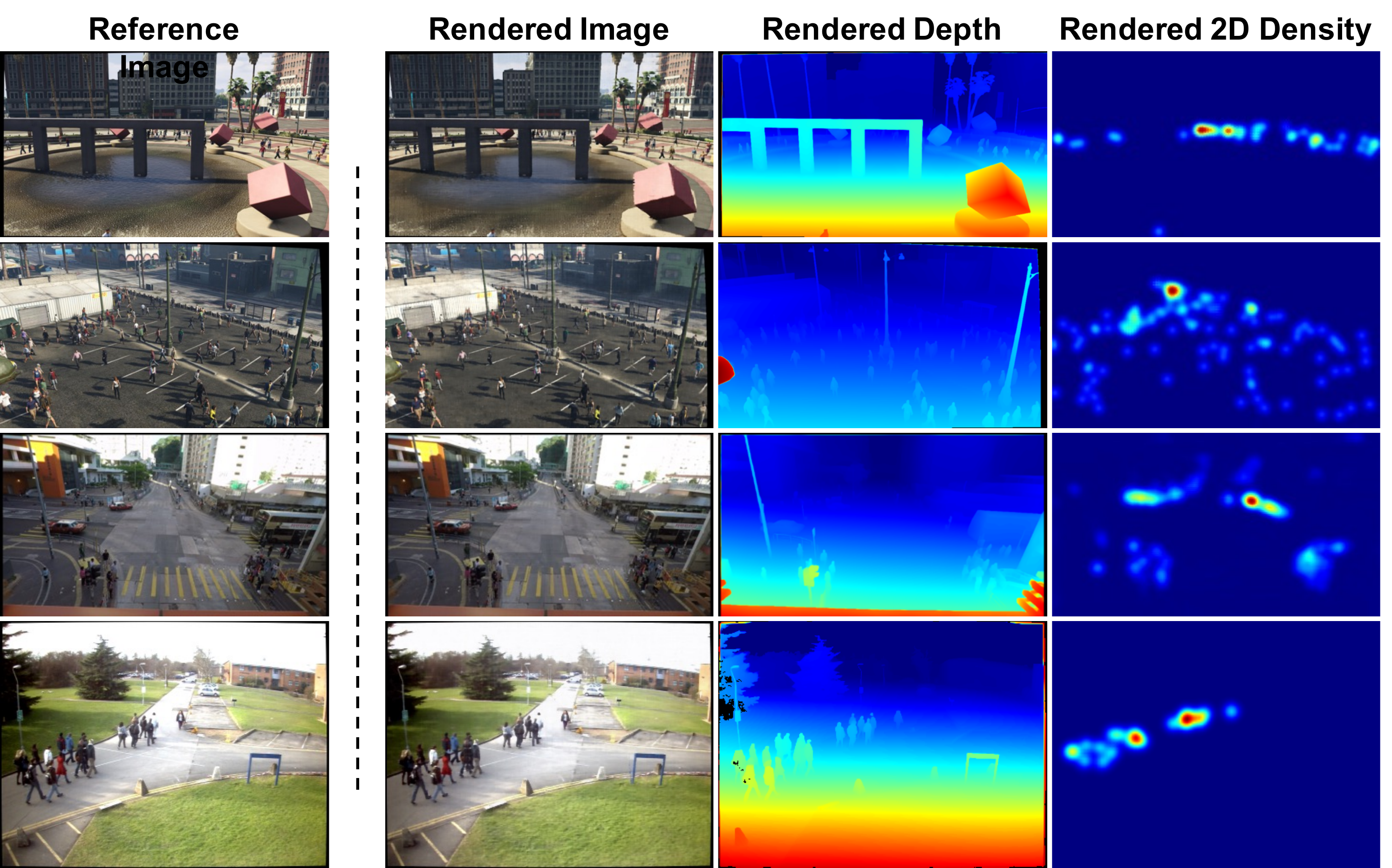} 
    \vspace{-2pt}
    \caption{
      {\bf Qualitative Results.} The figure illustrates several typical scenarios on the
      CVCS (1$_{\rm st}$~-~2$_{\rm nd}$ row), CityStreet (3$_{\rm rd}$ row) and PETS2009 (4$_{\rm th}$ row) datasets, including occlusion and congested crowds. 
      For each sample, the reference RGB image, the reconstructed RGB image, the rendered depth, and the projected 2D density map are presented, respectively.
    }
    \label{fig:decoder_render}
\end{figure}

Furthe, as Figure \ref{fig:decoder_render} visualizes the reconstructed RGB image, the rendered depth, and
the projected 2D density map, demonstrating that the SSLCounter is capable of reconstructing high-fidelity scene geometry and accurately capturing the spatial distribution of crowd density from MV images,
highlighting the effectiveness of the SSL volumetric rendering in enhancing both geometric and semantic understanding within the MVC framework.

\begin{table}
\centering
\resizebox{0.95\columnwidth}{!}
{%
\begin{tabular}{c |c c |c c}
\toprule
\multicolumn{1}{c|}{\multirow{2}{*}{\% Training -- Set}} & \multicolumn{2}{c|}{~~~~SSL~~~~} & \multicolumn{2}{c}{Performance}\\\cline{2-5}
& Training & Validation & MAE ($\downarrow$) & NAE ($\downarrow$)\\
\toprule
\multirow{2}{*}{70\%}   
& \no & \no &8.17& 0.072 \\
& \yes & \no & 5.02 & 0.043 \\
\hline
\multirow{3}{*}{100\%}   
& \no & \no &4.79& 0.039 \\
& \yes & \no & 4.01 & 0.032 \\
& \yes & \yes & 2.97 & 0.024 \\
\bottomrule
\end{tabular}
}

\caption{\textbf{Ablation Study on CVCS.} 
The table presents the ablation results on the CVCS \cite{zhang2021cross} dataset using scene-level MAE $\downarrow$ ane NAE $\downarrow$, where \% Training refers the proportion of data used for training, and SSL indicates applying the SSL on the corresponding data part.
}
\label{tb:ab_cvcs}
\end{table}

\subsection{Ablation Study}

This section presents ablation studies designed to quantify the contribution of individual components within the SSLCounter framework. 
The experiments validate that the  SSL strategy significantly addresses the annotation scarcity problem inherent in the MVC task. 
Notably, Tables \ref{tb:ab_citystreet} and \ref{tb:ab_cvcs} show that the framework delivers competitive performance while utilizing only 70\% of the labeled training data, underscoring its data efficiency and strong generalization capability. 
An in-depth evaluation of different SSL regularizers (Table \ref{tb:ssl_component}) further confirms that the multi-view projection consistency constraint is indispensable, which is a fundamental challenge in MVC

\begin{table}[htbp]
\centering
\resizebox{0.9\columnwidth}{!}
{%
\begin{tabular}{c |c c c |c c}
\toprule
\multicolumn{1}{c|}{\multirow{2}{*}{Benchmark}} & \multicolumn{3}{c|}{~~~~SSL~~~~} & \multicolumn{2}{c}{Performance}\\\cline{2-6}
& depth & density & color & MAE ($\downarrow$) & NAE ($\downarrow$)\\
\toprule

\multirow{4}{*}{CityStreet}   
& \no & \no & \no & 7.17 & 0.095 \\
& \yes & \no & \no & 6.71 & 0.092 \\
& \yes & \yes & \no &5.72 & 0.075 \\
& \yes & \yes & \yes &5.48 & 0.071 \\
\hline
\multirow{4}{*}{CVCS}  
& \no & \no & \no & 8.17 & 0.072 \\
& \yes & \no & \no &7.63 & 0.069 \\
& \yes & \yes & \no &5.80 & 0.051 \\
& \yes & \yes & \yes &5.02 & 0.043 \\
\bottomrule
\end{tabular}
}

\caption{\textbf{Ablation Study.}  
The table comprehensively ablates the impact of different regularizers that constitute the SSL using scene-level MAE $\downarrow$ and NAE $\downarrow$.
}
\label{tb:ssl_component}
\end{table}

\section{Conclusion}
\vspace{-2pt}
In this paper, we propose SSLCounter, a novel SSL paradigm for the MVC task that leverages volumetric neural rendering to alleviate the reliance on large-scale annotated datasets. 
Extensive experiments  demonstrate that SSLCounter achieves SOTA performance and superior data efficiency, maintaining competitive results even with reduced training data. 
Ablation studies further validate the effectiveness of the proposed SSL components. 
We believe SSLCounter provides a practical and robust solution for real-world MVC applications.


\clearpage
\section{Acknowledgement}
This research is supported by the Open Project Program of the State Key Laboratory of Virtual Reality Technology and Systems at Beihang University (Grant No. VRLAB2024C05). 

\bibliographystyle{IEEEbib}
\bibliography{refs}

\begin{thebibliography}{10}

\bibitem{zhang2019wide}
Qi~Zhang and Antoni~B Chan,
\newblock ``Wide-area crowd counting via ground-plane density maps and
  multi-view fusion cnns,''
\newblock in {\em Proceedings of the IEEE/CVF Conference on Computer Vision and
  Pattern Recognition (CVPR)}. IEEE, 2019, pp. 8297--8306.

\bibitem{zhang20203d}
Qi~Zhang and Antoni~B Chan,
\newblock ``3d crowd counting via multi-view fusion with 3d gaussian kernels,''
\newblock in {\em Proceedings of the AAAI Conference on Artificial Intelligence
  (AAAI)}, 2020, vol.~34, pp. 12837--12844.

\bibitem{zhang2021cross}
Qi~Zhang, Wei Lin, and Antoni~B Chan,
\newblock ``Cross-view cross-scene multi-view crowd counting,''
\newblock in {\em Proceedings of the IEEE/CVF Conference on Computer Vision and
  Pattern Recognition (CVPR)}. IEEE, 2021, pp. 557--567.

\bibitem{mo2024countformer}
Hong Mo, Xiong Zhang, Jianchao Tan, Cheng Yang, Qiong Gu, Bo~Hang, and Wenqi
  Ren,
\newblock ``Countformer: Multi-view crowd counting transformer,''
\newblock in {\em Proceedings of the European Conference on Computer Vision
  (ECCV)}. Springer, 2024, pp. 20--40.

\bibitem{zhang20223d}
Qi~Zhang and Antoni~B Chan,
\newblock ``3d crowd counting via geometric attention-guided multi-view
  fusion,''
\newblock {\em International Journal of Computer Vision (IJCV)}, vol. 130, no.
  12, pp. 3123--3139, 2022.

\bibitem{ristani2016performance}
Ergys Ristani, Francesco Solera, Roger Zou, Rita Cucchiara, and Carlo Tomasi,
\newblock ``Performance measures and a data set for multi-target, multi-camera
  tracking,''
\newblock in {\em Proceedings of the European Conference on Computer Vision
  (ECCV)}. Springer, 2016, pp. 17--35.

\bibitem{ferryman2009pets2009}
James Ferryman and Ali Shahrokni,
\newblock ``Pets2009: Dataset and challenge,''
\newblock in {\em IEEE International Workshop on Performance Evaluation of
  Tracking and Surveillance}. IEEE, 2009, pp. 1--6.

\bibitem{chibane2020neural}
Julian Chibane, Gerard Pons-Moll, et~al.,
\newblock ``Neural unsigned distance fields for implicit function learning,''
\newblock {\em Advances in Neural Information Processing Systems (NeurIPS)},
  vol. 33, pp. 21638--21652, 2020.

\bibitem{mildenhall2020nerf}
Ben Mildenhall, Pratul~P Srinivasan, Matthew Tancik, Jonathan~T Barron, Ravi
  Ramamoorthi, and Ren Ng,
\newblock ``Nerf: Representing scenes as neural radiance fields for view
  synthesis,''
\newblock in {\em Proceedings of the European Conference on Computer Vision
  (ECCV)}. Springer, 2020, pp. 405--421.

\bibitem{philion2020lift}
Jonah Philion and Sanja Fidler,
\newblock ``Lift, splat, shoot: Encoding images from arbitrary camera rigs by
  implicitly unprojecting to 3d,''
\newblock in {\em Proceedings of the European Conference on Computer Vision
  (ECCV)}. Springer, 2020, pp. 194--210.

\bibitem{liu2023sparsebev}
Haisong Liu, Yao Teng, Tao Lu, Haiguang Wang, and Limin Wang,
\newblock ``Sparsebev: High-performance sparse 3d object detection from
  multi-camera videos,''
\newblock in {\em Proceedings of the IEEE/CVF International Conference on
  Computer Vision (ICCV)}, 2023, pp. 18580--18590.

\bibitem{wang2021neus}
Peng Wang, Lingjie Liu, Yuan Liu, Christian Theobalt, Taku Komura, and Wenping
  Wang,
\newblock ``Neus: Learning neural implicit surfaces by volume rendering for
  multi-view reconstruction,''
\newblock {\em Advances in Neural Information Processing Systems}, vol. 34, pp.
  27171--27183, 2021.

\bibitem{oechsle2021unisurf}
Michael Oechsle, Songyou Peng, and Andreas Geiger,
\newblock ``Unisurf: Unifying neural implicit surfaces and radiance fields for
  multi-view reconstruction,''
\newblock in {\em Proceedings of the IEEE/CVF International Conference on
  Computer Vision (ICCV)}, 2021, pp. 5589--5599.

\bibitem{zhang2022calibration}
Qi~Zhang and Antoni~B Chan,
\newblock ``Calibration-free multi-view crowd counting,''
\newblock in {\em Proceedings of the European Conference on Computer Vision
  (ECCV)}. Springer, 2022, pp. 227--244.

\bibitem{zhang2022wide}
Qi~Zhang and Antoni~B Chan,
\newblock ``Wide-area crowd counting: Multi-view fusion networks for counting
  in large scenes,''
\newblock {\em International Journal of Computer Vision (IJCV)}, vol. 130, no.
  8, pp. 1938--1960, 2022.

\bibitem{zheng2021learning}
Liangfeng Zheng, Yongzhi Li, and Yadong Mu,
\newblock ``Learning factorized cross-view fusion for multi-view crowd
  counting,''
\newblock in {\em Proceedings of the IEEE International Conference on
  Multimedia and Expo (ICME)}. IEEE, 2021, pp. 1--6.

\bibitem{zhai2022co}
Qiang Zhai, Fan Yang, Xin Li, Guo-Sen Xie, Hong Cheng, and Zicheng Liu,
\newblock ``Co-communication graph convolutional network for multi-view crowd
  counting,''
\newblock {\em IEEE Transactions on Multimedia (TMM)}, vol. 25, pp. 5813--5825,
  2022.

\bibitem{qiu2019cross}
Haibo Qiu, Chunyu Wang, Jingdong Wang, Naiyan Wang, and Wenjun Zeng,
\newblock ``Cross view fusion for 3d human pose estimation,''
\newblock in {\em Proceedings of the IEEE/CVF International Conference on
  Computer Vision (ICCV)}. IEEE, 2019, pp. 4342--4351.

\end{thebibliography}
\end{document}